\documentclass{article} 
\usepackage{nips13submit_e,times}
\usepackage{hyperref}
\usepackage{url}
\usepackage{microtype}
\usepackage{graphicx}
\usepackage{subfigure}
\usepackage{amsmath}
\usepackage{setspace}
\def\ie{{\em i.e.}}
\def\eg{{\em e.g.}}
\def\etal{{\em et al.}}

\newcommand{\ignore}[1]{}

\nipsfinalcopy 

\begin{document}

\title{Accurate Face Detection for High Performance}
\author{
  Faen Zhang, Xinyu Fan, Guo Ai, Jianfei Song, Yongqiang Qin, Jiahong Wu\\
  AInnovation Technology Ltd, Beijing, China\\
  \texttt{\{zhangfaen,fanxinyu,aiguo\}@ainnovation.com}\\
  \texttt{\{songjianfei,qinyongqiang,wujiahong\}@ainnovation.com}\\
}
\maketitle

\begin{abstract}
Face detection has witnessed significant progress due to the advances of deep convolutional neural networks (CNNs). Its central issue in recent years is how to improve the detection performance of tiny faces. To this end, many recent works propose some specific strategies, redesign the architecture and introduce new loss functions for tiny object detection. In this report, we start from the popular one-stage RetinaNet~\cite{DBLP:conf/iccv/LinPRK17} approach and apply some recent tricks to obtain a high performance face detector namely AInnoFace. Specifically, we apply the Intersection over Union (IoU) loss function~\cite{DBLP:conf/mm/YuJWCH16} for regression, employ the two-step classification and regression~\cite{DBLP:journals/corr/abs-1809-02693} for detection, revisit the data augmentation based on data-anchor-sampling~\cite{tang2018pyramidbox} for training, utilize the max-out operation~\cite{DBLP:conf/iccv/abs-1708-05237} for classification and use the multi-scale testing strategy~\cite{DBLP:conf/iccv/abs-1708-05237} for inference. As a consequence, the proposed face detection method achieves state-of-the-art performance on the most popular and challenging face detection benchmark WIDER FACE~\cite{DBLP:conf/cvpr/YangLLT16} dataset.
\end{abstract}

\section{Introduction}
Face detection is a tremendously important field in computer vision needed for face recognition, sentiment analysis, video surveillance, and many other fields. Given an arbitrary image, the goal of face detection is to determine whether there are any faces in the image, and if present, return the image location and extent of each face. The recent issue of face detection is how to improve the detection performance in unrestricted scenarios. Because detecting faces in real-world images has many difficulties including occlusion, significant scale variation, different illumination conditions, various facial poses, rich facial expressions, etc. Many works are devoted to solving this issue and great progress has been achieved with the development of deep convolutional neural networks (CNNs). For example, the average precision (AP) performance on the challenging WIDER FACE dataset~\cite{DBLP:conf/cvpr/YangLLT16} has been improved from $40\%$ to $90\%$ over recent years.

To improve the performance of face detection in unrestricted scenarios where exists plenty of tiny faces, some works~\cite{DBLP:conf/icb/YangYLL14,DBLP:conf/iccv/YangLLT15,DBLP:journals/spl/ZhangZLQ16,DBLP:conf/icpr/Ohn-BarT16a,DBLP:journals/corr/YangXLT17} combine traditional methods (\eg, cascade-based mechanism and part-based in DPM) with deep learning methods (\eg, CNN) to perform face detection. A number of works~\cite{DBLP:journals/corr/ZhuZLS16,DBLP:conf/cvpr/HuR17,DBLP:conf/iccv/NajibiSCD17} resort to the context information around the face region to find tiny faces based on the Faster R-CNN~\cite{DBLP:journals/pami/RenHG017} and SSD~\cite{DBLP:conf/eccv/LiuAESRFB16} detectors. Several works~\cite{DBLP:conf/eccv/CaiFFV16,DBLP:journals/corr/abs-1809-02693,DBLP:journals/corr/abs-1712-00721,DBLP:journals/corr/abs-1811-08557} redesign the architecture of modern object detection to better detect tiny faces. A series of works~\cite{DBLP:journals/corr/WangLJW17,DBLP:journals/corr/abs-1709-05256,DBLP:conf/iccv/abs-1708-05237,zhu2018seeing,DBLP:journals/corr/abs-1802-02142} propose some special strategies for tiny faces into the generic object detection methods to improve face detection performance. There are many works~\cite{tang2018pyramidbox,DBLP:journals/corr/abs-1901-06651,DBLP:journals/corr/abs-1810-10220} present some new data augmentations for tiny faces to improve the performance. Some works~\cite{wang2017fan,DBLP:journals/corr/abs-1901-02350} introduce the attention mechanism on the feature maps to focus on face regions for better detection performance.

In this work, we first modify the popular one-stage RetinaNet~\cite{DBLP:conf/iccv/LinPRK17} method to perform face detection as our baseline model. Then some recent tricks are applied on this baseline to develop a high performance face detector namely AInnoFace: (1) Employing the two-step classification and regression for detection; (2) Applying the Intersection over Union (IoU) loss function for regression; (3) Revisiting the data augmentation based on data-anchor-sampling for training; (4) Utilizing the max-out operation for robuster classification; (5) Using the multi-scale testing strategy for inference. Consequently, we achieve some new state-of-the-art AP results on the challenging face detection benchmark WIDER FACE~\cite{DBLP:conf/cvpr/YangLLT16} dataset.

\section{Related Work}
\subsection{Traditional Method}
Face detection has been extensively studied from its emergence in the 1990s to the present because of its wide practical applications. The pioneering work~\cite{DBLP:journals/ijcv/ViolaJ04} of Viola and Jones uses the Haar-like feature and the AdaBoost strategy to train several cascaded face detectors, achieving a very good tradeoff between accuracy and efficiency in some simple and fixed scenarios. Afterwards, subsequent works~\cite{DBLP:journals/ijcv/BrubakerWSMR08,DBLP:journals/pami/LiaoJL16} have made great progress by developing more advanced features and more powerful classifiers. Apart from the boosted cascade methods, several studies~\cite{DBLP:conf/cvpr/YanLWL14,DBLP:conf/cvpr/ZhuR12,DBLP:conf/icb/YangYLL14} introduce another famous framework of Deformable Part Model (DPM)~\cite{DBLP:conf/eccv/MathiasBPG14} to the filed of face detection task, which detect faces by modelling the relationship of deformable facial parts and achieve promising performance in some simple application scenarios. However, these traditional face detectors are unreliable in complex scenarios because they depend on non-robust hand-crafted features and classifiers.

\subsection{Deep Learning Method}
Deep learning approaches significantly boost the recent progress in the face detection filed and the CNN-based face detectors have achieved the highest performance in the last few years. The cascade CNN-based methods~\cite{DBLP:conf/cvpr/LiLSBH15,DBLP:conf/cvpr/QinYLH16} train a series of CNN models separately or jointly to perform face detection, and achieve promising accuracy and efficiency simultaneously. After that, MTCNN~\cite{DBLP:journals/spl/ZhangZLQ16} and PCN~\cite{shi2018real} add another extra branch to detect five facial landmarks and predict face angles via the multi-task learning in a coarse-to-fine manner under a cascade-style structure. Faceness~\cite{DBLP:conf/iccv/YangLLT15} obtains different scores according to the spatial structure and arrangement of facial parts to detect faces under severe occlusion and unconstrained pose variations. LDCF+~\cite{DBLP:conf/icpr/Ohn-BarT16a} utilizes the boosted decision tree classifier to detect faces. UnitBox~\cite{DBLP:conf/mm/YuJWCH16} introduces an Intersection-over-Union (IoU) loss to directly minimize the IoUs of the predictions and the ground-truths for more accurate location. ScaleFace~\cite{DBLP:journals/corr/YangXLT17} detects different scales of faces via applying a specialized set of CNNs with different structures. SAFD~\cite{hao2017scale} develops a scale proposal stage to automatically normalize face sizes prior to detection. Hu~\etal~\cite{DBLP:conf/cvpr/HuR17} explore the contextual information with some separate detectors for different scales to find tiny faces. S$^{2}$AP~\cite{song2018beyond} finds face via paying attention to specific scales in image pyramid and valid locations in each scales layer. Zhu~\etal~\cite{zhu2018seeing} use the Expected Max Overlapping (EMO) score to evaluate the quality of anchor setting. Bai~\etal~\cite{bai2018finding} generate a clear super-resolution face from a blurry small one via to GAN~\cite{goodfellow2014generative} to detect blurry small faces.

Besides, many state-of-the-are face detectors are evolved from generic object detection methods including the two-stage approach (Faster R-CNN~\cite{DBLP:journals/pami/RenHG017}, R-FCN~\cite{DBLP:conf/nips/DaiLHS16} and FPN~\cite{DBLP:conf/cvpr/LinDGHHB17}) and the one-stage approach (SSD~\cite{DBLP:conf/eccv/LiuAESRFB16}, RefineDet~\cite{DBLP:journals/corr/abs-1711-06897} and RetinaNet~\cite{DBLP:conf/iccv/LinPRK17}). Based on Faster R-CNN and R-FCN, some face detection methods (\eg, Face R-CNN~\cite{DBLP:journals/corr/WangLJW17}, Face R-FCN~\cite{DBLP:journals/corr/abs-1709-05256}, CMS-RCNN~\cite{DBLP:journals/corr/ZhuZLS16} and FDNet~\cite{DBLP:journals/corr/abs-1802-02142}) design several specific training and testing strategies with the consideration of the characteristics of face detection. Similar to FPN, FANet~\cite{DBLP:journals/corr/abs-1712-00721} aggregates higher-level features to augment lower-level features for face detection. Based on SSD, DCFPN~\cite{DBLP:conf/ccbr/ZhangZLSWL17} and FaceBoxes~\cite{DBLP:conf/ijcb/abs-1708-05234} design a lightweight face detection network to achieve CPU real-time speed with promising result. In contrast, high performance face detectors including S$^{3}$FD~\cite{DBLP:conf/iccv/abs-1708-05237}, SFDet~\cite{zhang2019single}, SSH~\cite{DBLP:conf/iccv/NajibiSCD17}, PyramidBox~\cite{tang2018pyramidbox,li2019pyramidbox++} and DSFD~\cite{DBLP:journals/corr/abs-1810-10220} equip SSD with some specific strategies for better detection of small faces, such as architecture diagram, training strategy, contextual reasoning and multiple layers exploiting. Besides, FAN~\cite{wang2017fan} and DFS~\cite{DBLP:journals/corr/abs-1811-08557} use different types of attention mechanism on RetinaNet to handle hard faces. SRN~\cite{DBLP:journals/corr/abs-1809-02693} combines the multi-step detection in RefineDet~\cite{DBLP:journals/corr/abs-1711-06897} and the focal loss in RetinaNet~\cite{DBLP:conf/iccv/LinPRK17} to perform efficient and accurate face detection. After that, VIM-FD~\cite{DBLP:journals/corr/abs-1901-02350} and ISRN~\cite{DBLP:journals/corr/abs-1901-06651} combine many previous techniques on SRN and achieve new state-of-the-art performance.

\begin{figure*}[t!]
\centering
\includegraphics[width=1.0\linewidth]{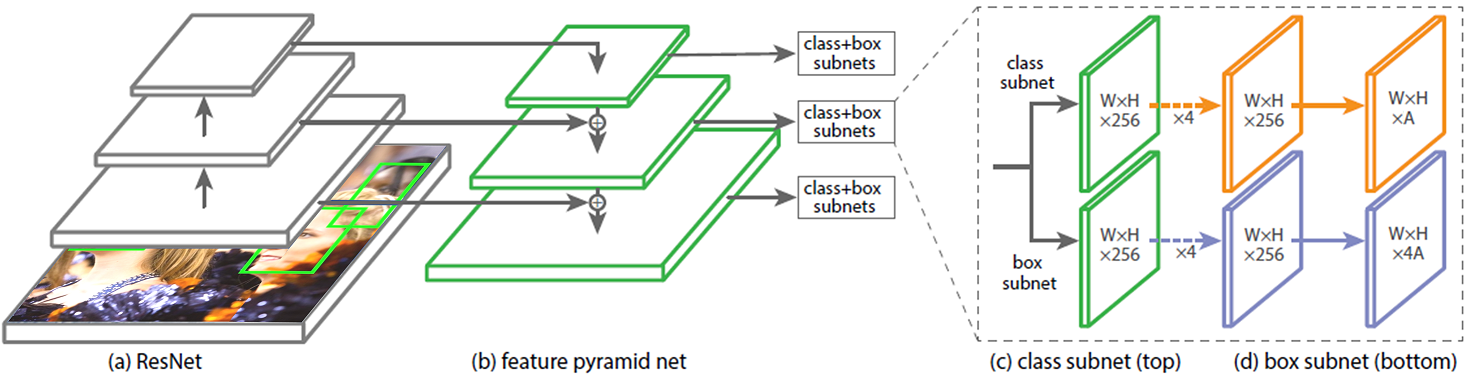}
\caption{Architecture of our face detector baseline built on RetinaNet~\cite{DBLP:conf/iccv/LinPRK17}. (a) Backbone: a feed-forward ResNet-152~\cite{DBLP:conf/cvpr/HeZRS16} architecture extracts the multi-scale feature maps. (b) Neck: a 6-level Feature Pyramid Network (FPN)~\cite{DBLP:conf/cvpr/LinDGHHB17} structure generates a richer multi-scale convolutional feature pyramid. After that, two shared subnetworks are attached, one for classifying anchor boxes (c) and one for regressing from anchor boxes to ground-truth object boxes (d). At the end, we use the focal loss~\cite{DBLP:conf/iccv/LinPRK17} for the binary classification and the IoU loss~\cite{DBLP:conf/mm/YuJWCH16} for the regression.}
\label{fig:framework}
\end{figure*}

\section{Method}
Starting from the RetinaNet~\cite{DBLP:conf/iccv/LinPRK17} face detector baseline, we apply some recently proposed strategies to achieve state-of-the-art performance on the challenging WIDER FACE~\cite{DBLP:conf/cvpr/YangLLT16} dataset.

\subsection{RetinaNet Baseline}
RetinaNet is one of the most popular one-stage object detection methods. One-stage detectors that are applied over a regular, dense sampling of possible object locations have the potential to be faster and simpler, but have trailed the accuracy of two-stage detectors because of extreme class imbalance encountered during training. To solve this issue, the focal loss is proposed in RetinaNet as follow:
\begin{equation}
FL(p_t) = -\alpha_t(1 - p_t)^\gamma \log (p_t).
\end{equation}
and
\begin{equation}
p_t =
\begin{cases}
p &\text{if $y = 1$}\\
1 - p &\text{otherwise}
\end{cases}
\end{equation}
where $y \in \{\pm1\}$ specifies the ground-truth class, $p \in [0,1]$ is the model's estimated probability for the class with label $y=1$, $\alpha_t$ is a balanced factor and $\gamma$ a tunable focusing parameter. The focal loss is the reshaping of cross entropy loss such that it down-weights the loss assigned to well-classified examples. The novel focal loss focuses training on a sparse set of hard examples and prevents the vast number of easy negatives from overwhelming the detector during training.

In this report, we take the modified RetinaNet shown in Figure~\ref{fig:framework} as our baseline, which is a single and unified network composed of a backbone network, a neck network and two task-specific subnetworks. The backbone and neck networks are responsible for computing a multi-scale convolutional feature maps over an entire input image. The first subnet performs classification on the backbones output and the second subnet performs convolution bounding box regression. Specifically, we adopt the ResNet-152~\cite{DBLP:conf/cvpr/HeZRS16} with 6-level feature pyramid structure as our backbone network. We follow the way in FPN~\cite{DBLP:conf/cvpr/LinDGHHB17} to generate the 6-level feature maps (P2 to P7) for detection.

\subsection{IoU Regression Loss}
The object detection task consists of the classification subtask and the regression subtask. For regression, the smooth L1 loss~\cite{DBLP:conf/iccv/Girshick15} is the common loss function used to reduce the difference between anchor boxes and ground-truth bounding boxes, and the Intersection over Union (IoU) is the most popular evaluation metric used in the object detection benchmarks. However, as indicated in~\cite{rezatofighi2019generalized}, there is a gap between optimizing the commonly used smooth L1 distance losses for regressing the parameters of a bounding box and maximizing this IoU metric value. The optimal objective for a metric should be the metric itself, so we follow the UnitBox~\cite{DBLP:conf/mm/YuJWCH16} to minimize the difference between the predictions and the ground-truths via directly using IoU as the regression loss.

The IoU regression loss function is defined as below:
\begin{equation}
\begin{aligned}
{L_{IoU}}= -\ln \frac{Intersection(B_{p}, B_{gt})}{Union(B_{p}, B_{gt})}
\end{aligned}
\end{equation}
where $B_{p}=(x_1,y_1,x_2,y_2)$ and $B_{gt}=(x^*_1,y^*_1,x^*_2,y^*_2)$ are the predicted bounding box and the ground-truth bounding box respectively, $Intersection()$ and $Union()$ indicate the intersection and union area between $B_{p}$ and $B_{gt}$.

\subsection{Selective Refinement Network}
As described in Selective Refinement Network (SRN)~\cite{DBLP:journals/corr/abs-1809-02693}, using RetinaNet to perform face detection still exists two problems: (a) low recall efficiency: the precision is not high enough at high recall rates, \ie, the Precision-Recall curve extends far enough to the right but not steep enough; (b) low location accuracy: the performance drops dramatically as the IoU threshold increases, \ie, the accuracy of the bounding box location needs to be improved. To solve the aforementioned two issues, Selective Two-step Classification (STC) and Selective Two-step Regression (STR) are proposed in SRN and we follow these designs to further improve the performance of our face detector.

STC conducts two-step classification on three low level detection layers to filter out most simple negatives and reduce the search space for the subsequent classifier. Its loss function is:
\begin{equation}
\begin{aligned}
L_{STC} (\{p_i\},\{q_i\})=\frac{1}{N_{{s}_1}} \sum_{i\in \Omega}L_{FL}(p_i,l_i^\ast) + \frac{1}{N_{{s}_2}} \sum_{i\in \Phi}L_{FL}(q_i, l_i^\ast),
\end{aligned}
\end{equation}
STR performs is two-step regression on three high level detection layers to adjust anchors and provide better initialization for the subsequent regressor. Its loss function is: 
\begin{equation}
\begin{aligned}
L_{STR}(\{x_i\},\{t_i\})=\frac{1}{N_{{s}_1}} \sum_{i\in \Psi}[l_i^\ast=1]L_{r}(x_i, g_i^\ast) + \frac{1}{N_{{s}_2}} \sum_{i\in \Phi}[l_i^\ast=1]{L}_{r}(t_i, g_i^\ast),
\end{aligned}
\end{equation}
where $i$ is the anchor index, $p_i$/$q_i$ and $x_i$/$t_i$ are the prediction of classification and regression in the first/second step, $l_i^\ast$/$g_i^\ast$ are the ground truth of class/location, $N_{s_{1}}$/$N_{s_{2}}$ are the number of positive anchors in the first/second step, $\Omega$/$\Psi$ are the collection of classification/regression samples for the first step and $\Phi$ is the collection of samples for the second step, $L_{FL}$ is the sigmoid focal loss and $[l_i^\ast=1]{\cal L}_{\text{r}}$ indicates that the IoU regression loss is computed only for positive anchors.

\subsection{Data Augmentation}
Date augmentation is of importance for one-stage detectors to construct a robust model to adapt to variations of objects, especially for face detection where has plenty tiny faces. Following most of the face detectors, we randomly expand and crop the original training images with additional random photometric distortion~\cite{DBLP:journals/corr/Howard13} and flipping to generate the training samples. Besides, with probability of $0.5$, we replace the above random cropping operation with the anchor-based sampling like data-anchor-sampling in PyramidBox~\cite{tang2018pyramidbox} to diversify the scale distribution of training samples. The anchor-based sampling operation first randomly selects a face of size $S_{face}$ in a batch, then finds its nearest anchor scale $S_{anchor}$. After that, it chooses a random scale $S_{random}$ around the nearest anchor scale. Finally, it resizes the image by $S^* = S_{random} / S_{face}$ and randomly crops a standard size of the training size containing the selected face to get the anchor-sampled training data.

\subsection{Max-out Label}
To reduce the tiny false positives from background regions, the max-out operation for the background class is introduced in~\cite{DBLP:conf/iccv/abs-1708-05237}, and then the following work~\cite{tang2018pyramidbox} uses this max-out operation on both foreground and background classes. In the proposed face detection method, the max-out operation is applied in the classification subnet to recall more faces and reduce false positives simultaneously. To be more specific, the classification subnet first predicts $c_p + c_n$ scores for each anchor, and then selects $\max\{c_p\}$ and $\max\{c_n\}$ as the final face and no-face confidence score to compute the classification loss. Empirically, we set $c_p = 3$ and $c_n = 3$ to train our final model.

\subsection{Multi-scale Testing}
Since there are plenty of tiny faces in the challenging WIDER FACE dataset, the multi-scale testing strategy is useful to improve the performance. We use the open source code\footnote{\url{https://github.com/sfzhang15/SFD/blob/master/sfd_test_code/WIDER_FACE/wider_test.py}} to conduct the multi-scale testing during inference. It inputs the image to the trained model multiple times with different sizes and then merge these detection results with the bounding box voting operation.

\section{Experiment}

\subsection{Experimental Dataset}
\begin{figure}[!h]
\centering
\includegraphics[width=1.0\textwidth]{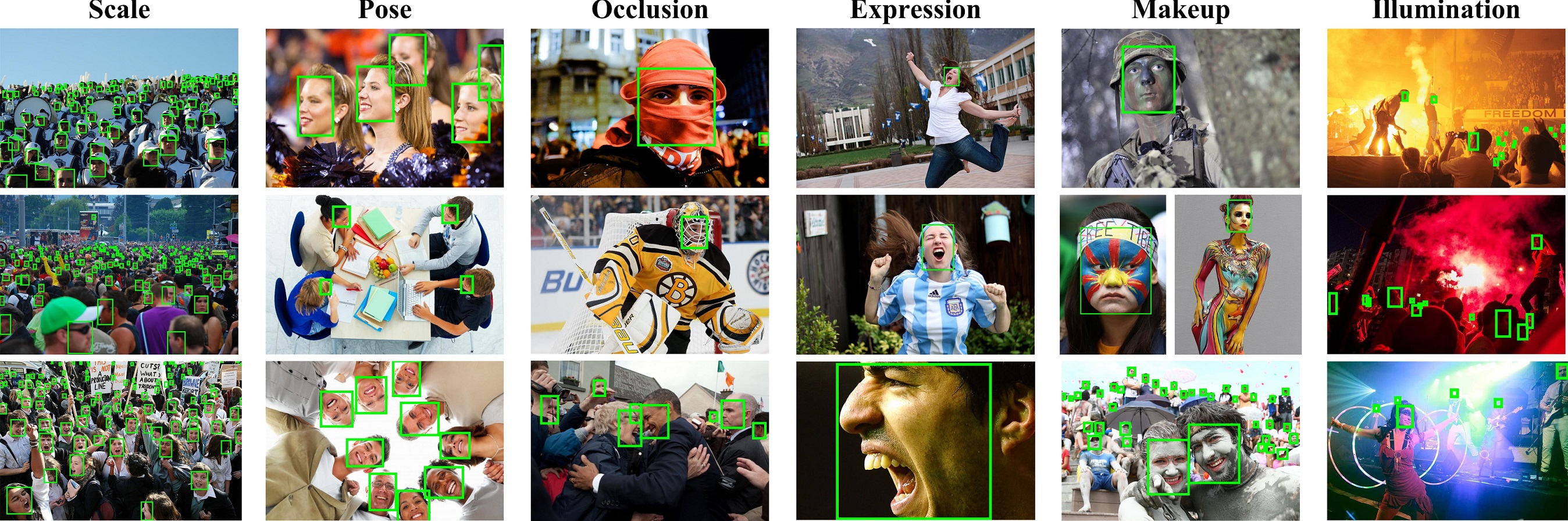}
\caption{Some sample images for each attribute of WIDER FACE~\cite{DBLP:conf/cvpr/YangLLT16}.}
\label{fig:widerface}
\end{figure}

We verify the proposed AInnoFace detector on the WIDER FACE~\cite{DBLP:conf/cvpr/YangLLT16} dataset, which is a popular face detection benchmark dataset and whose images are selected from the publicly available WIDER~\cite{xiong2015wider} dataset. The WIDER FACE dataset contains $32,203$ images and $393,703$ annotated face bounding boxes with a high degree of variability in scale, pose, occlusion, expression, makeup and illumination as depicted in Figure~\ref{fig:widerface}. All the images are organized based on $61$ event classes and are randomly selected from each event class by $40\%$/$10\%$/$50\%$ as training, validation and testing subsets. Based on the detection rate of EdgeBox~\cite{DBLP:conf/eccv/ZitnickD14}, the validation and testing subsets are divided into three difficulty levels: Easy, Medium, Hard. The Average Precision (AP) is adopted as the evaluation metric. Following MALF~\cite{DBLP:conf/fgr/YangYLL15} and Caltech~\cite{dollar2012pedestrian} datasets, this dataset does not release bounding box ground truth for the testing images and researchers are required to submit final prediction files to get the AP performance on the testing subset. The proposed method is trained on the training subset and evaluated on both validation and testing subsets.

\subsection{Anchor Detail}
Following \cite{DBLP:journals/corr/abs-1809-02693} we set two $2S$ and $2\sqrt{2}S$ anchor scales ($S$ is the downsampling factor of detection layer) and one $1.25$ aspect ratio. Thus, there are $A=2$ anchors at each location, covering the scale of $8-362$ pixels in the $1024\times1024$ input image. During the training phase, anchors are assigned to ground-truth boxes using the $\theta_{p}$ IoU threshold and to background if their IoU is in $[0, \theta_{n})$. The rest anchors in $[\theta_{n}, \theta_{p})$ are ignored. We set $\theta_{n}=0.3$ and $\theta_{p}=0.7$ for the first step, and $\theta_{n}=0.4$ and $\theta_{p}=0.5$ for the second step.

\begin{figure}[!h]
\centering
\subfigure[Val: Easy]{
\label{fig:ve}
\includegraphics[width=0.49\linewidth]{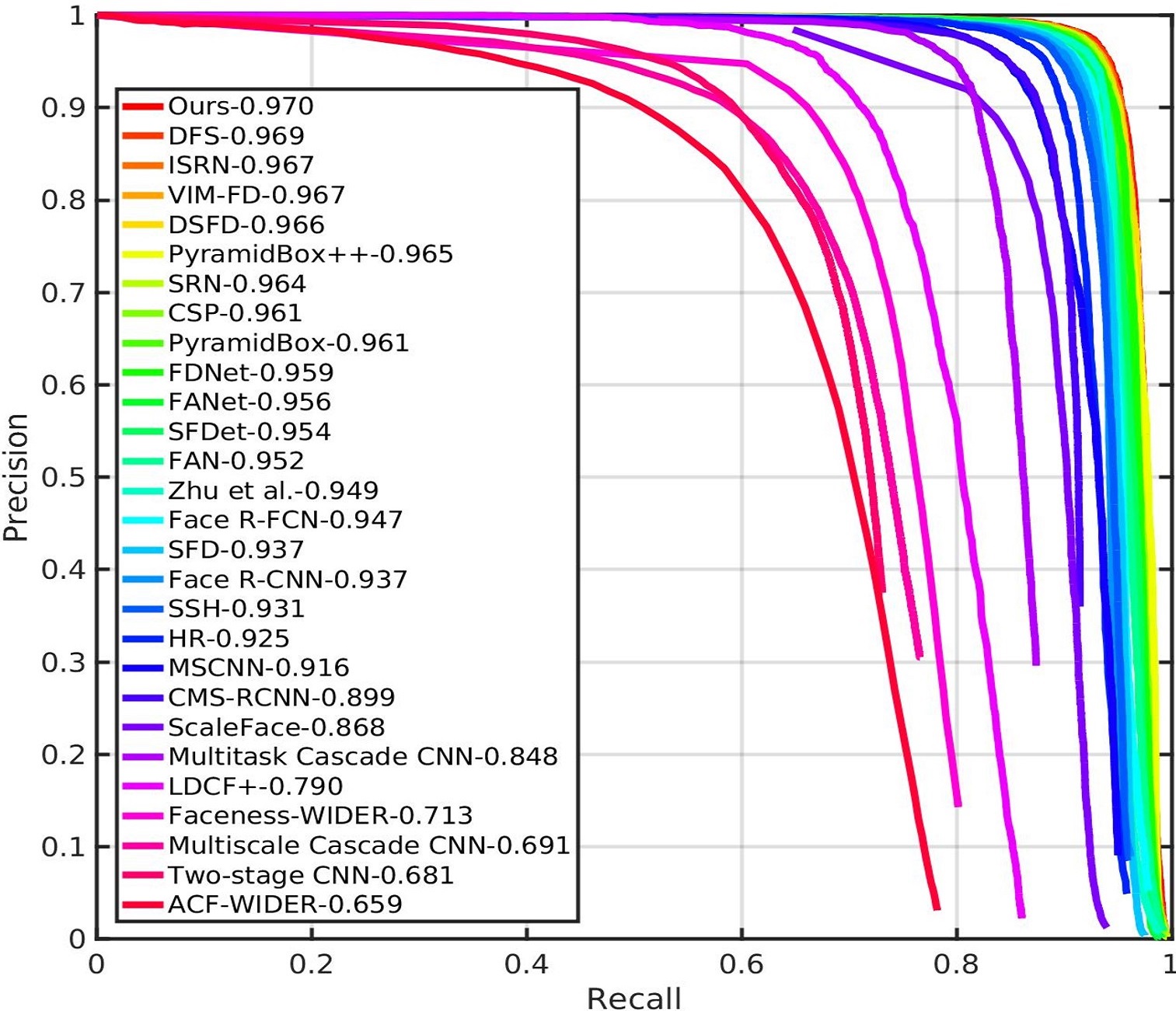}}
\subfigure[Test: Easy]{
\label{fig:te}
\includegraphics[width=0.49\linewidth]{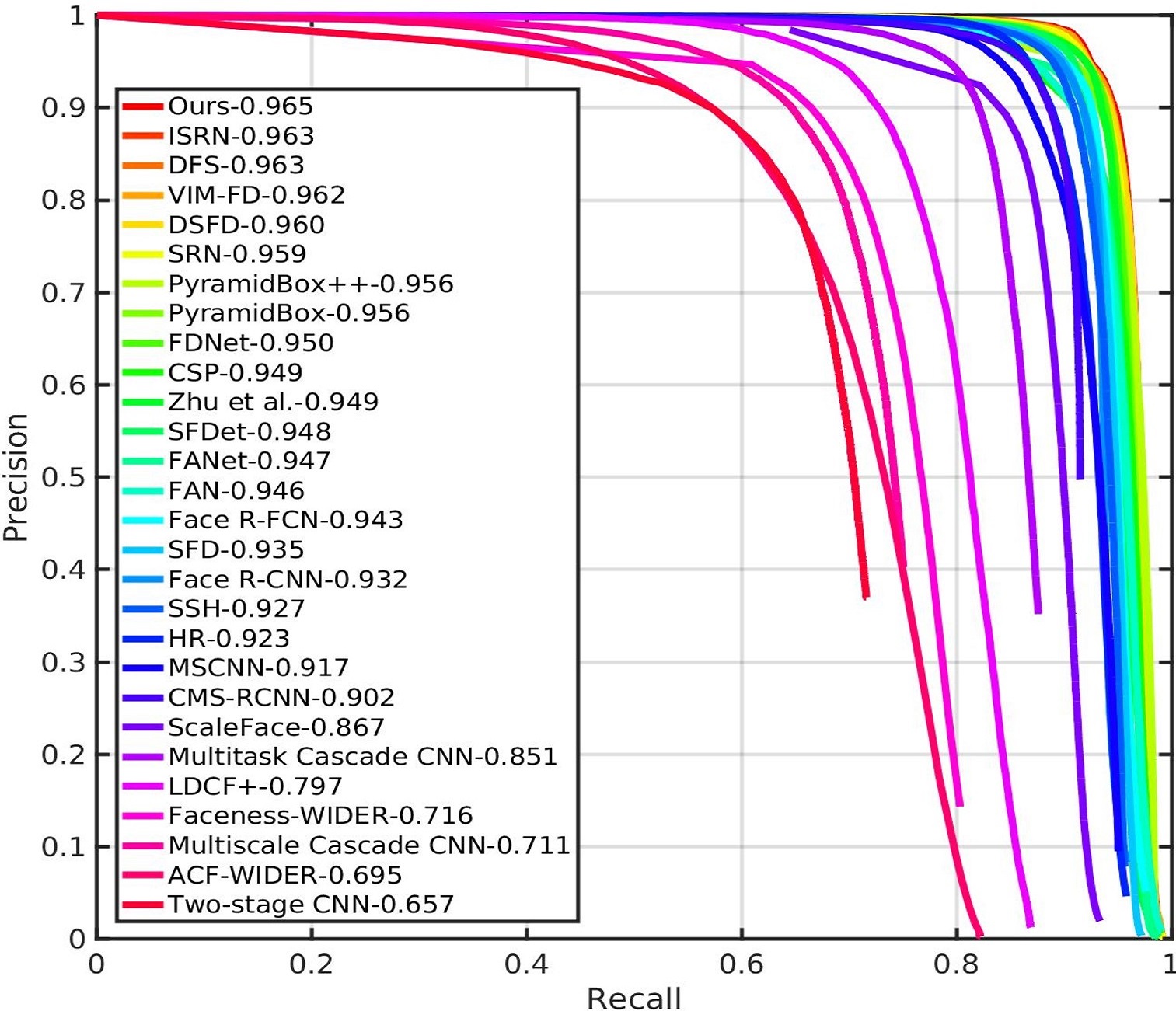}}
\subfigure[Val: Medium]{
\label{fig:vm}
\includegraphics[width=0.49\linewidth]{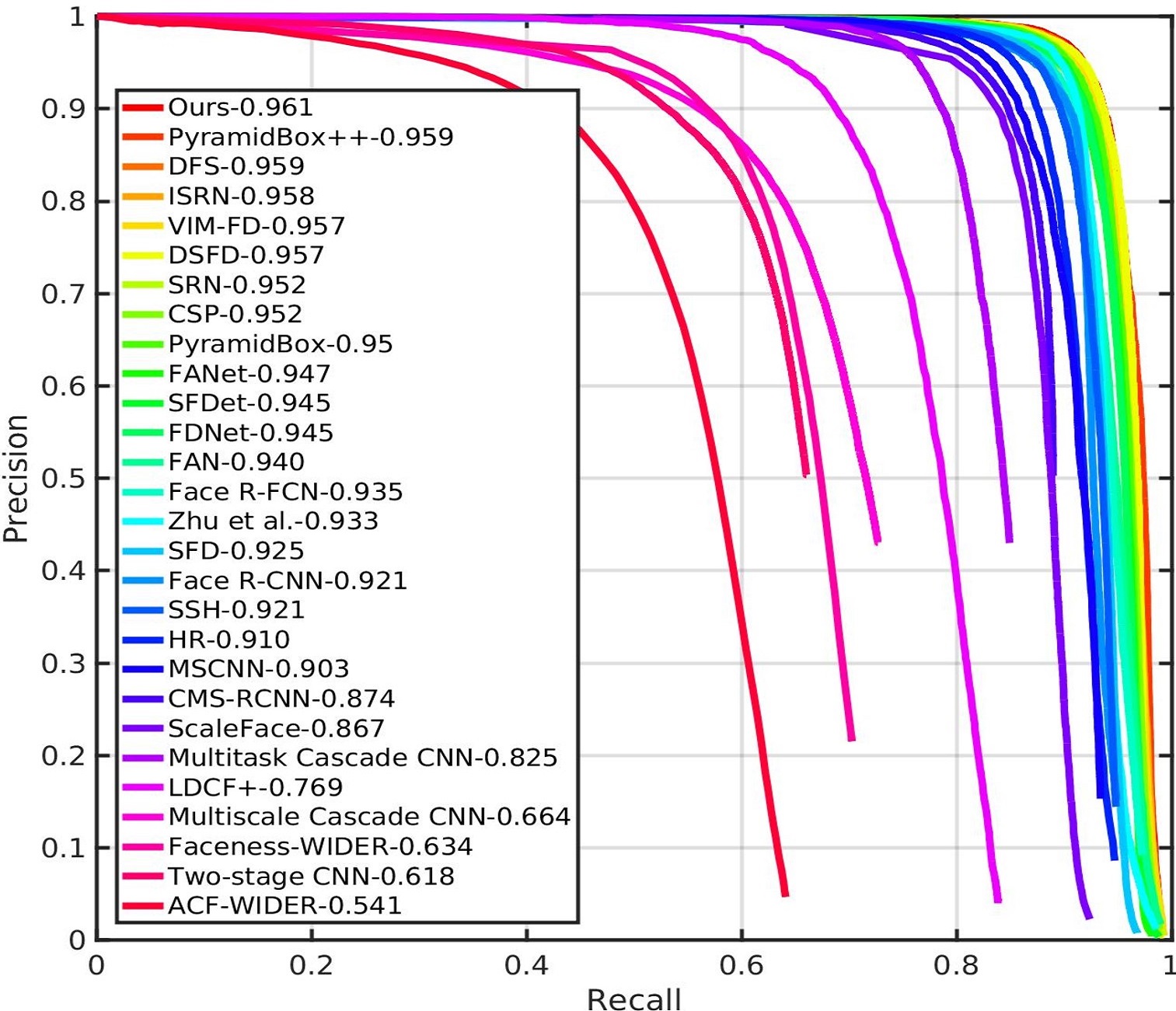}}
\subfigure[Test: Medium]{
\label{fig:tm}
\includegraphics[width=0.49\linewidth]{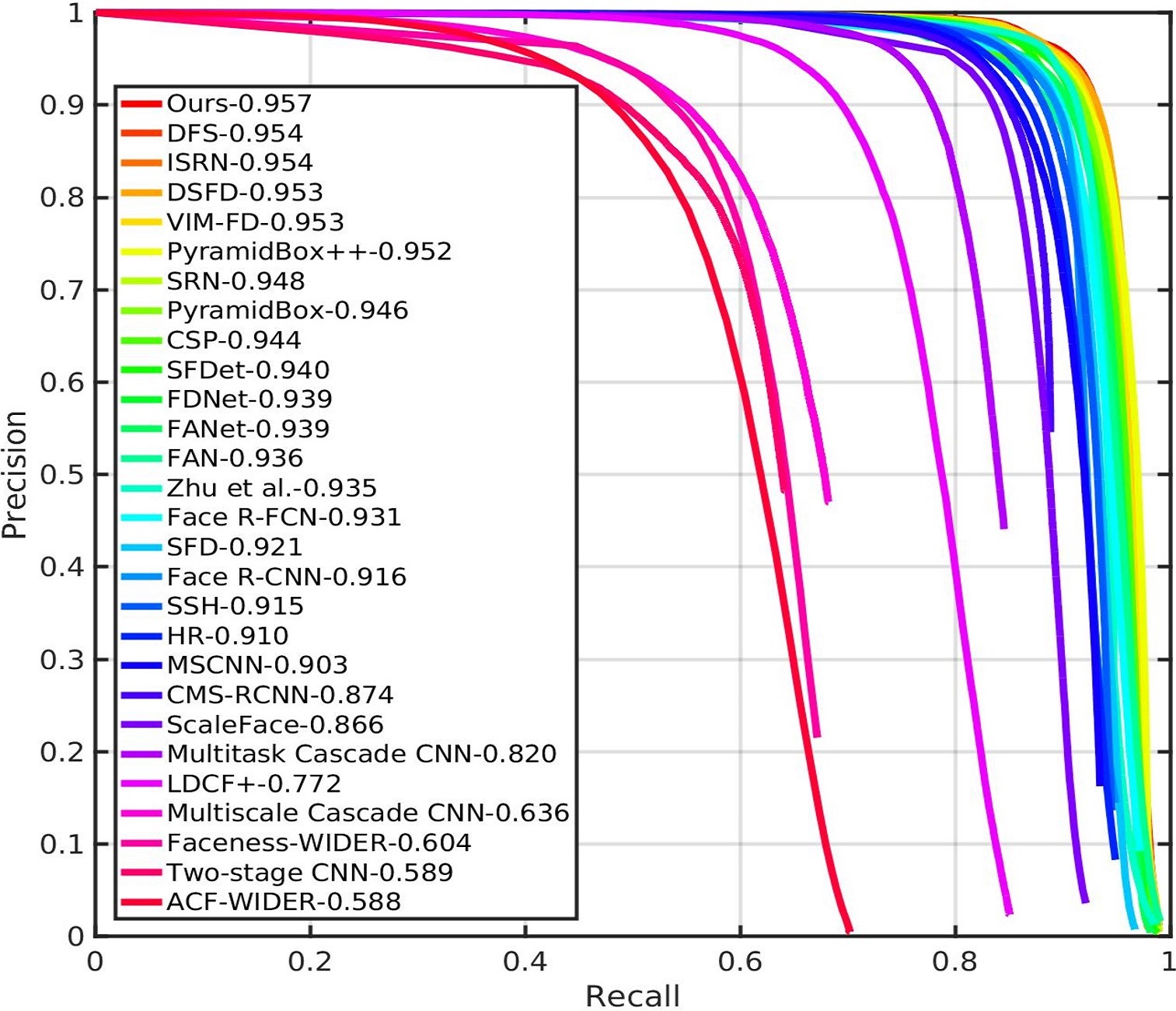}}
\subfigure[Val: Hard]{
\label{fig:vh}
\includegraphics[width=0.49\linewidth]{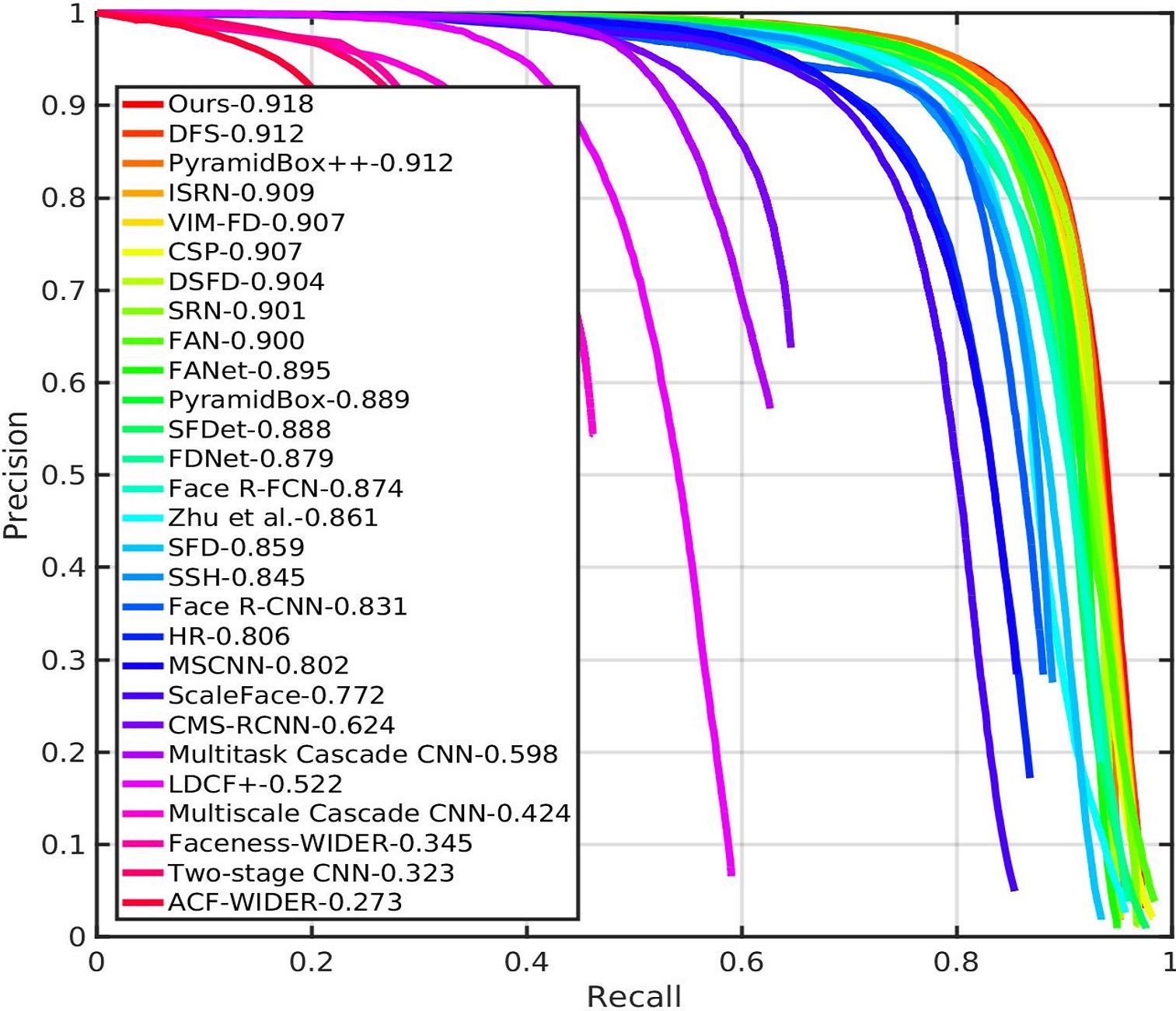}}
\subfigure[Test: Hard]{
\label{fig:th}
\includegraphics[width=0.49\linewidth]{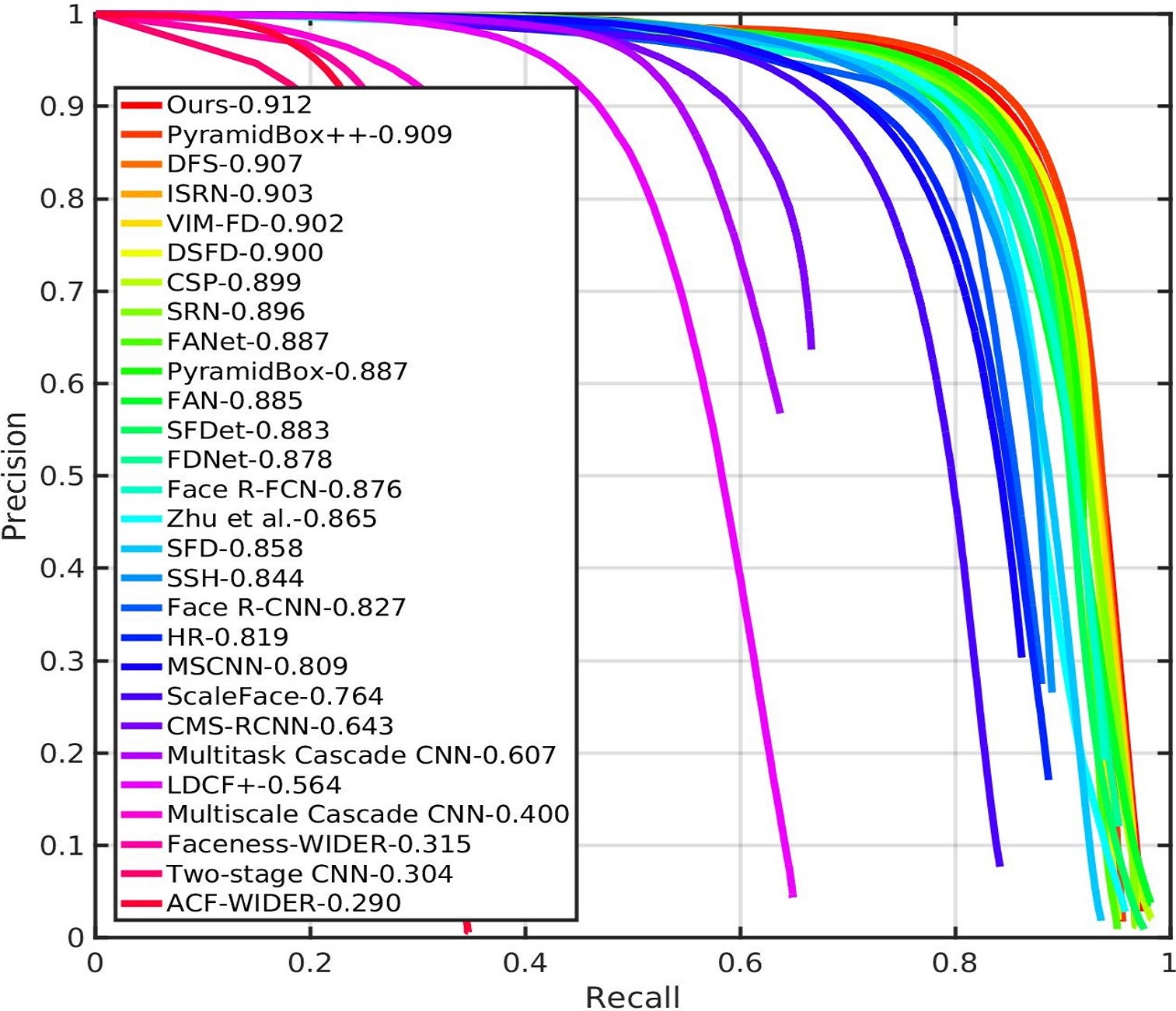}}
\caption{Precision-recall curves on WIDER FACE validation and testing subsets.}
\label{fig:wider-face-ap}
\end{figure}

\subsection{Optimization Detail}
The backbone network in the proposed AInnoFace detector is initialized by the pretrained model on the ImageNet~\cite{DBLP:journals/ijcv/RussakovskyDSKS15} dataset. We use the ``xavier''~\cite{DBLP:journals/jmlr/GlorotB10} method to randomly initialize the parameters in the newly added convolutional layers. The stochastic gradient descent (SGD) algorithm is used to fine-tune the model with $0.9$ momentum, $0.0001$ weight decay and batch size $32$. The warmup strategy~\cite{DBLP:journals/corr/GoyalDGNWKTJH17} is applied to gradually ramp up the learning rate from $0.0003125$ to $0.01$ at the first $5$ epochs. After that, it switches to the regular learning rate schedule, i.e., dividing by $10$ at $100$ and $120$ epochs and ending at $130$ epochs. The full training and testing codes are built on the PyTorch library~\cite{paszke2017pytorch}.

\subsection{Evaluation Result}
Figure~\ref{fig:wider-face-ap} shows the comparison of the proposed AInnoFace detector with twenty-seven state-of-the-art methods~\cite{DBLP:conf/eccv/CaiFFV16,DBLP:journals/corr/abs-1809-02693,DBLP:conf/cvpr/HuR17,DBLP:journals/corr/abs-1810-10220,DBLP:conf/iccv/NajibiSCD17,DBLP:conf/icpr/Ohn-BarT16a,tang2018pyramidbox,DBLP:journals/corr/abs-1811-08557,DBLP:journals/corr/WangLJW17,wang2017fan,DBLP:journals/corr/abs-1709-05256,DBLP:conf/icb/YangYLL14,DBLP:conf/iccv/YangLLT15,DBLP:conf/cvpr/YangLLT16,DBLP:journals/corr/YangXLT17,DBLP:journals/corr/abs-1802-02142,DBLP:journals/corr/abs-1712-00721,DBLP:journals/spl/ZhangZLQ16,DBLP:conf/iccv/abs-1708-05237,zhu2018seeing,DBLP:journals/corr/ZhuZLS16,DBLP:journals/corr/abs-1901-02350,DBLP:journals/corr/abs-1901-06651,li2019pyramidbox++,zhang2019single} on both the validation and testing subsets based on the precision-recall curve and AP. As shown in Figure~\ref{fig:wider-face-ap}, our face detector sets some new state-of-the-art results based on the AP score across the three subsets on both validation and testing subsets, \ie, $97.0\%$ (Easy), $96.1\%$ (Medium) and $91.8\%$ (Hard) for validation subset, and $96.5\%$ (Easy), $95.7\%$ (Medium) and $91.2\%$ (Hard) for testing subset. These results outperform all the compared state-of-the-art methods and demonstrate the superiority of our AInnoFace detector.

\section{Conclusion}
In this report, we present a high performance face detector by equipping the popular one-stage RetinaNet~\cite{DBLP:conf/iccv/LinPRK17} method with some recent tricks: (1) Employing the two-step classification and regression for detection; (2) Applying the Intersection over Union (IoU) loss function for regression; (3) Revisiting the data augmentation based on data-anchor-sampling for training; (4) Utilizing the max-out operation for robuster classification; (5) Using the multi-scale testing strategy for inference. Experiments on the WIDER FACE dataset demonstrate that the proposed AInnoFace detector achieves the state-of-the-art detection performance.

\bibliographystyle{plain}
\bibliography{reference}

\begin{thebibliography}{10}

\bibitem{bai2018finding}
Yancheng Bai, Yongqiang Zhang, Mingli Ding, and Bernard Ghanem.
\newblock Finding tiny faces in the wild with generative adversarial network.
\newblock In {\em CVPR}, 2018.

\bibitem{DBLP:journals/ijcv/BrubakerWSMR08}
S.~Charles Brubaker, Jianxin Wu, Jie Sun, Matthew~D. Mullin, and James~M. Rehg.
\newblock On the design of cascades of boosted ensembles for face detection.
\newblock {\em IJCV}, 2008.

\bibitem{DBLP:conf/eccv/CaiFFV16}
Zhaowei Cai, Quanfu Fan, Rog{\'{e}}rio~Schmidt Feris, and Nuno Vasconcelos.
\newblock A unified multi-scale deep convolutional neural network for fast
  object detection.
\newblock In {\em ECCV}, 2016.

\bibitem{DBLP:journals/corr/abs-1809-02693}
Cheng Chi, Shifeng Zhang, Junliang Xing, Zhen Lei, Stan~Z. Li, and Xudong Zou.
\newblock Selective refinement network for high performance face detection.
\newblock In {\em AAAI}, 2019.

\bibitem{DBLP:conf/nips/DaiLHS16}
Jifeng Dai, Yi~Li, Kaiming He, and Jian Sun.
\newblock {R-FCN:} object detection via region-based fully convolutional
  networks.
\newblock In {\em NIPS}, 2016.

\bibitem{dollar2012pedestrian}
Piotr Dollar, Christian Wojek, Bernt Schiele, and Pietro Perona.
\newblock Pedestrian detection: An evaluation of the state of the art.
\newblock {\em TPAMI}, 2012.

\bibitem{DBLP:conf/iccv/Girshick15}
Ross~B. Girshick.
\newblock Fast {R-CNN}.
\newblock In {\em ICCV}, 2015.

\bibitem{DBLP:journals/jmlr/GlorotB10}
Xavier Glorot and Yoshua Bengio.
\newblock Understanding the difficulty of training deep feedforward neural
  networks.
\newblock In {\em AISTATS}, 2010.

\bibitem{goodfellow2014generative}
Ian Goodfellow, Jean Pouget-Abadie, Mehdi Mirza, Bing Xu, David Warde-Farley,
  Sherjil Ozair, Aaron Courville, and Yoshua Bengio.
\newblock Generative adversarial nets.
\newblock In {\em NIPS}, 2014.

\bibitem{DBLP:journals/corr/GoyalDGNWKTJH17}
Priya Goyal, Piotr Doll{\'{a}}r, Ross~B. Girshick, Pieter Noordhuis, Lukasz
  Wesolowski, Aapo Kyrola, Andrew Tulloch, Yangqing Jia, and Kaiming He.
\newblock Accurate, large minibatch {SGD:} training imagenet in 1 hour.
\newblock {\em CoRR}, 2017.

\bibitem{hao2017scale}
Zekun Hao, Yu~Liu, Hongwei Qin, Junjie Yan, Xiu Li, and Xiaolin Hu.
\newblock Scale-aware face detection.
\newblock In {\em CVPR}, 2017.

\bibitem{DBLP:conf/cvpr/HeZRS16}
Kaiming He, Xiangyu Zhang, Shaoqing Ren, and Jian Sun.
\newblock Deep residual learning for image recognition.
\newblock In {\em CVPR}, 2016.

\bibitem{DBLP:journals/corr/Howard13}
Andrew~G. Howard.
\newblock Some improvements on deep convolutional neural network based image
  classification.
\newblock {\em CoRR}, 2013.

\bibitem{DBLP:conf/cvpr/HuR17}
Peiyun Hu and Deva Ramanan.
\newblock Finding tiny faces.
\newblock In {\em CVPR}, 2017.

\bibitem{DBLP:conf/cvpr/LiLSBH15}
Haoxiang Li, Zhe Lin, Xiaohui Shen, Jonathan Brandt, and Gang Hua.
\newblock A convolutional neural network cascade for face detection.
\newblock In {\em CVPR}, 2015.

\bibitem{DBLP:journals/corr/abs-1810-10220}
Jian Li, Yabiao Wang, Changan Wang, Ying Tai, Jianjun Qian, Jian Yang, Chengjie
  Wang, Ji{-}Lin Li, and Feiyue Huang.
\newblock {DSFD:} dual shot face detector.
\newblock In {\em CVPR}, 2019.

\bibitem{li2019pyramidbox++}
Zhihang Li, Xu~Tang, Junyu Han, Jingtuo Liu, and Ran He.
\newblock Pyramidbox++: High performance detector for finding tiny face.
\newblock {\em CoRR}, 2019.

\bibitem{DBLP:journals/pami/LiaoJL16}
Shengcai Liao, Anil~K. Jain, and Stan~Z. Li.
\newblock A fast and accurate unconstrained face detector.
\newblock {\em TPAMI}, 2016.

\bibitem{DBLP:conf/cvpr/LinDGHHB17}
Tsung{-}Yi Lin, Piotr Doll{\'{a}}r, Ross~B. Girshick, Kaiming He, Bharath
  Hariharan, and Serge~J. Belongie.
\newblock Feature pyramid networks for object detection.
\newblock In {\em CVPR}, 2017.

\bibitem{DBLP:conf/iccv/LinPRK17}
Tsung{-}Yi Lin, Priya Goyal, Ross~B. Girshick, Kaiming He, and Piotr
  Doll{\'{a}}r.
\newblock Focal loss for dense object detection.
\newblock In {\em ICCV}, 2017.

\bibitem{DBLP:conf/eccv/LiuAESRFB16}
Wei Liu, Dragomir Anguelov, Dumitru Erhan, Christian Szegedy, Scott~E. Reed,
  Cheng{-}Yang Fu, and Alexander~C. Berg.
\newblock {SSD:} single shot multibox detector.
\newblock In {\em ECCV}, 2016.

\bibitem{DBLP:conf/eccv/MathiasBPG14}
Markus Mathias, Rodrigo Benenson, Marco Pedersoli, and Luc J.~Van Gool.
\newblock Face detection without bells and whistles.
\newblock In {\em ECCV}, 2014.

\bibitem{DBLP:conf/iccv/NajibiSCD17}
Mahyar Najibi, Pouya Samangouei, Rama Chellappa, and Larry~S. Davis.
\newblock {SSH:} single stage headless face detector.
\newblock In {\em ICCV}, 2017.

\bibitem{DBLP:conf/icpr/Ohn-BarT16a}
Eshed Ohn{-}Bar and Mohan~M. Trivedi.
\newblock To boost or not to boost? on the limits of boosted trees for object
  detection.
\newblock In {\em ICPR}, 2016.

\bibitem{paszke2017pytorch}
Adam Paszke, Sam Gross, Soumith Chintala, and Gregory Chanan.
\newblock Pytorch, 2017.

\bibitem{DBLP:conf/cvpr/QinYLH16}
Hongwei Qin, Junjie Yan, Xiu Li, and Xiaolin Hu.
\newblock Joint training of cascaded {CNN} for face detection.
\newblock In {\em CVPR}, 2016.

\bibitem{DBLP:journals/pami/RenHG017}
Shaoqing Ren, Kaiming He, Ross~B. Girshick, and Jian Sun.
\newblock Faster {R-CNN:} towards real-time object detection with region
  proposal networks.
\newblock {\em TPAMI}, 2017.

\bibitem{rezatofighi2019generalized}
Hamid Rezatofighi, Nathan Tsoi, JunYoung Gwak, Amir Sadeghian, Ian Reid, and
  Silvio Savarese.
\newblock Generalized intersection over union: A metric and a loss for bounding
  box regression.
\newblock In {\em CVPR}, 2019.

\bibitem{DBLP:journals/ijcv/RussakovskyDSKS15}
Olga Russakovsky, Jia Deng, Hao Su, Jonathan Krause, Sanjeev Satheesh, Sean Ma,
  Zhiheng Huang, Andrej Karpathy, Aditya Khosla, Michael~S. Bernstein,
  Alexander~C. Berg, and Fei{-}Fei Li.
\newblock Imagenet large scale visual recognition challenge.
\newblock {\em IJCV}, 2015.

\bibitem{shi2018real}
Xuepeng Shi, Shiguang Shan, Meina Kan, Shuzhe Wu, and Xilin Chen.
\newblock Real-time rotation-invariant face detection with progressive
  calibration networks.
\newblock In {\em CVPR}, 2018.

\bibitem{song2018beyond}
Guanglu Song, Yu~Liu, Ming Jiang, Yujie Wang, Junjie Yan, and Biao Leng.
\newblock Beyond trade-off: Accelerate fcn-based face detector with higher
  accuracy.
\newblock In {\em CVPR}, 2018.

\bibitem{tang2018pyramidbox}
Xu~Tang, Daniel~K Du, Zeqiang He, and Jingtuo Liu.
\newblock Pyramidbox: A context-assisted single shot face detector.
\newblock In {\em ECCV}, 2018.

\bibitem{DBLP:journals/corr/abs-1811-08557}
Wanxin Tian, Zixuan Wang, Haifeng Shen, Weihong Deng, Binghui Chen, and Xiubao
  Zhang.
\newblock Learning better features for face detection with feature fusion and
  segmentation supervision.
\newblock {\em CoRR}, 2018.

\bibitem{DBLP:journals/ijcv/ViolaJ04}
Paul~A. Viola and Michael~J. Jones.
\newblock Robust real-time face detection.
\newblock {\em IJCV}, 2004.

\bibitem{DBLP:journals/corr/WangLJW17}
Hao Wang, Zhifeng Li, Xing Ji, and Yitong Wang.
\newblock Face {R-CNN}.
\newblock {\em CoRR}, 2017.

\bibitem{wang2017fan}
Jianfeng Wang, Ye~Yuan, and Gang Yu.
\newblock Face attention network: An effective face detector for the occluded
  faces.
\newblock {\em CoRR}, 2017.

\bibitem{DBLP:journals/corr/abs-1709-05256}
Yitong Wang, Xing Ji, Zheng Zhou, Hao Wang, and Zhifeng Li.
\newblock Detecting faces using region-based fully convolutional networks.
\newblock {\em CoRR}, 2017.

\bibitem{xiong2015wider}
Yuanjun Xiong, Kai Zhu, Dahua Lin, and Xiaoou Tang.
\newblock Recognize complex events from static images by fusing deep channels.
\newblock In {\em CVPR}, 2015.

\bibitem{DBLP:conf/cvpr/YanLWL14}
Junjie Yan, Zhen Lei, Longyin Wen, and Stan~Z. Li.
\newblock The fastest deformable part model for object detection.
\newblock In {\em CVPR}, 2014.

\bibitem{DBLP:conf/icb/YangYLL14}
Bin Yang, Junjie Yan, Zhen Lei, and Stan~Z. Li.
\newblock Aggregate channel features for multi-view face detection.
\newblock In {\em IJCB}, 2014.

\bibitem{DBLP:conf/fgr/YangYLL15}
Bin Yang, Junjie Yan, Zhen Lei, and Stan~Z. Li.
\newblock Fine-grained evaluation on face detection in the wild.
\newblock In {\em FG}, 2015.

\bibitem{DBLP:conf/iccv/YangLLT15}
Shuo Yang, Ping Luo, Chen~Change Loy, and Xiaoou Tang.
\newblock From facial parts responses to face detection: {A} deep learning
  approach.
\newblock In {\em ICCV}, 2015.

\bibitem{DBLP:conf/cvpr/YangLLT16}
Shuo Yang, Ping Luo, Chen~Change Loy, and Xiaoou Tang.
\newblock {WIDER} {FACE:} {A} face detection benchmark.
\newblock In {\em CVPR}, 2016.

\bibitem{DBLP:journals/corr/YangXLT17}
Shuo Yang, Yuanjun Xiong, Chen~Change Loy, and Xiaoou Tang.
\newblock Face detection through scale-friendly deep convolutional networks.
\newblock {\em CoRR}, 2017.

\bibitem{DBLP:conf/mm/YuJWCH16}
Jiahui Yu, Yuning Jiang, Zhangyang Wang, Zhimin Cao, and Thomas~S. Huang.
\newblock Unitbox: An advanced object detection network.
\newblock In {\em ACMMM}, 2016.

\bibitem{DBLP:journals/corr/abs-1802-02142}
Changzheng Zhang, Xiang Xu, and Dandan Tu.
\newblock Face detection using improved faster {RCNN}.
\newblock {\em CoRR}, 2018.

\bibitem{DBLP:journals/corr/abs-1712-00721}
Jialiang Zhang, Xiongwei Wu, Jianke Zhu, and Steven C.~H. Hoi.
\newblock Feature agglomeration networks for single stage face detection.
\newblock {\em CoRR}, 2017.

\bibitem{DBLP:journals/spl/ZhangZLQ16}
Kaipeng Zhang, Zhanpeng Zhang, Zhifeng Li, and Yu~Qiao.
\newblock Joint face detection and alignment using multitask cascaded
  convolutional networks.
\newblock {\em SPL}, 2016.

\bibitem{DBLP:journals/corr/abs-1711-06897}
Shifeng Zhang, Longyin Wen, Xiao Bian, Zhen Lei, and Stan~Z. Li.
\newblock Single-shot refinement neural network for object detection.
\newblock In {\em CVPR}, 2018.

\bibitem{zhang2019single}
Shifeng Zhang, Longyin Wen, Hailin Shi, Zhen Lei, Siwei Lyu, and Stan~Z Li.
\newblock Single-shot scale-aware network for real-time face detection.
\newblock {\em IJCV}, 2019.

\bibitem{DBLP:journals/corr/abs-1901-06651}
Shifeng Zhang, Rui Zhu, Xiaobo Wang, Hailin Shi, Tianyu Fu, Shuo Wang, Tao Mei,
  and Stan~Z. Li.
\newblock Improved selective refinement network for face detection.
\newblock {\em CoRR}, 2019.

\bibitem{DBLP:conf/ccbr/ZhangZLSWL17}
Shifeng Zhang, Xiangyu Zhu, Zhen Lei, Hailin Shi, Xiaobo Wang, and Stan~Z. Li.
\newblock Detecting face with densely connected face proposal network.
\newblock In {\em CCBR}, 2017.

\bibitem{DBLP:conf/ijcb/abs-1708-05234}
Shifeng Zhang, Xiangyu Zhu, Zhen Lei, Hailin Shi, Xiaobo Wang, and Stan~Z. Li.
\newblock Faceboxes: {A} {CPU} real-time face detector with high accuracy.
\newblock In {\em IJCB}, 2017.

\bibitem{DBLP:conf/iccv/abs-1708-05237}
Shifeng Zhang, Xiangyu Zhu, Zhen Lei, Hailin Shi, Xiaobo Wang, and Stan~Z. Li.
\newblock S\({}^{\mbox{3}}\){FD}: Single shot scale-invariant face detector.
\newblock In {\em ICCV}, 2017.

\bibitem{DBLP:journals/corr/abs-1901-02350}
Yundong Zhang, Xiang Xu, and Xiaotao Liu.
\newblock Robust and high performance face detector.
\newblock {\em CoRR}, 2019.

\bibitem{zhu2018seeing}
Chenchen Zhu, Ran Tao, Khoa Luu, and Marios Savvides.
\newblock Seeing small faces from robust anchor’s perspective.
\newblock In {\em CVPR}, 2018.

\bibitem{DBLP:journals/corr/ZhuZLS16}
Chenchen Zhu, Yutong Zheng, Khoa Luu, and Marios Savvides.
\newblock {CMS-RCNN:} contextual multi-scale region-based {CNN} for
  unconstrained face detection.
\newblock {\em CoRR}, 2016.

\bibitem{DBLP:conf/cvpr/ZhuR12}
Xiangxin Zhu and Deva Ramanan.
\newblock Face detection, pose estimation, and landmark localization in the
  wild.
\newblock In {\em CVPR}, 2012.

\bibitem{DBLP:conf/eccv/ZitnickD14}
C.~Lawrence Zitnick and Piotr Doll{\'{a}}r.
\newblock Edge boxes: Locating object proposals from edges.
\newblock In {\em ECCV}, 2014.

\end{thebibliography}

\end{document}